# CNN-Based Joint Clustering and Representation Learning with Feature Drift Compensation for Large-Scale Image Data

Chih-Chung Hsu, *Member*, *IEEE* and Chia-Wen Lin, *Senior Member*, *IEEE*

*Abstract*—Given a large unlabeled set of images, how to efficiently and effectively group them into clusters based on extracted visual representations remains a challenging problem. To address this problem, we propose a convolutional neural network (CNN) to jointly solve clustering and representation learning in an iterative manner. In the proposed method, given an input image set, we first randomly pick $k$ samples and extract their features as initial cluster centroids using the proposed CNN with an initial model pre-trained from the ImageNet dataset. Mini-batch $k$-means is then performed to assign cluster labels to individual input samples for a mini-batch of images randomly sampled from the input image set until all images are processed. Subsequently, the proposed CNN simultaneously updates the parameters of the proposed CNN and the centroids of image clusters iteratively based on stochastic gradient descent. We also propose a feature drift compensation scheme to mitigate the drift error caused by feature mismatch in representation learning. Experimental results demonstrate the proposed method outperforms start-of-the-art clustering schemes in terms of accuracy and storage complexity on large-scale image sets containing millions of images.

## I. INTRODUCTION

IMAGE clustering [1]–[16] is a fundamental problem for many image processing and computer vision applications. Nowadays, a huge number of images have been uploaded to cloud for sharing or storage. How to efficiently organize such large-scale image data is a challenging issue. In general, most research works on large-scale image clustering were based on feature encoding, such as hashing [17][18], which can largely reduce the dimensionality of image features so as to make large-scale clustering possible. However, reducing the dimensionality of features is equivalent to decreasing the

Manuscript received December 20, 2016, revised April 2 2017 and July 14 2017, accepted August 1 2017. Date of publication August 2, 2017; date of current version Month Day, 2017. This work was supported in part by the Ministry of Science and Technology, Taiwan, under Grants MOST 103-2221-E-007-046-MY3 and MOST 105-2218-E-007-012. The guest editor coordinating the review of this manuscript and approving it for publication was Dr. Jingkuan Song.
Chih-Chung Hsu is with the Institute of Communication Engineering, National Tsing Hua University, Hsinchu, Taiwan.
Chia-Wen Lin (corresponding author) is with the Department of Electrical Engineering and the Institute of Communications Engineering, National Tsing Hua University, Hsinchu, Taiwan. (e-mail: cwlin@ee.nthu.edu.tw).

representational power, leading to unsatisfactory clustering performance. Besides, the hash-based approaches usually assume that features are extracted before hash encoding. Different feature representations might lead to redesigning hash functions because of different number of dimensions or value ranges of feature vectors [1].

Clustering methods can be roughly categorized into hierarchical clustering and centroid-based clustering. The most popular algorithms for hierarchical clustering are agglomerative clustering [3], [4]. In agglomerative clustering, initially, individual samples in input data are considered as a cluster containing a single sample. Then, in each iteration, the two closest clusters in the raw or feature domain are merged into a cluster and the centroid of the merged cluster is computed accordingly. By iteratively merging the two closest clusters and updating the associated cluster centroids each time, we finally obtain the desired number of clusters and the corresponding centroids, which is, however, computationally very expensive for a large dataset.

In contrast, centroid-based clustering (e.g., k-means and spectral clustering) [6]–[11] randomly picks $k$ samples from the input data as initial cluster centroids. Then, each sample finds its closest cluster centroid and is assigned with the corresponding cluster label. As a result, the cluster centroids are updated according to the clustering result. The clustering and centroid updating are sequentially iterated until converging to a solution point. To further improve clustering performance, some advanced techniques such as spectral clustering and matrix factorization [6] were proposed to map visual features to another discriminative feature space to boost clustering performance. Such centroid-based clustering is more suitable for large-scale data clustering than hierarchical clustering due to less memory usage and computational power requirements. The effectiveness of centroid-based clustering, nevertheless, highly relies on feature representational power.

### A. Deep-Learning Based Image Clustering

Recently, deep-learning-based approaches have reached a series of breakthroughs in various fields such as image classification [19], object detection and tracking, and retrieval [20]. The most popular network architectures for image/video applications including AlexNet [19], ResNet [21] VGG [22], Inception module [23], and FCN [24] are all based on convolutional neural networks (CNNs). CNNs have proven to be able to learn significantly better discriminative visual representations for images compared to traditional hand-crafted features or features learned by shallow neural networks, given a sufficiently large labeled training set (usually containing millions of images like those provided in ImageNet [25]).

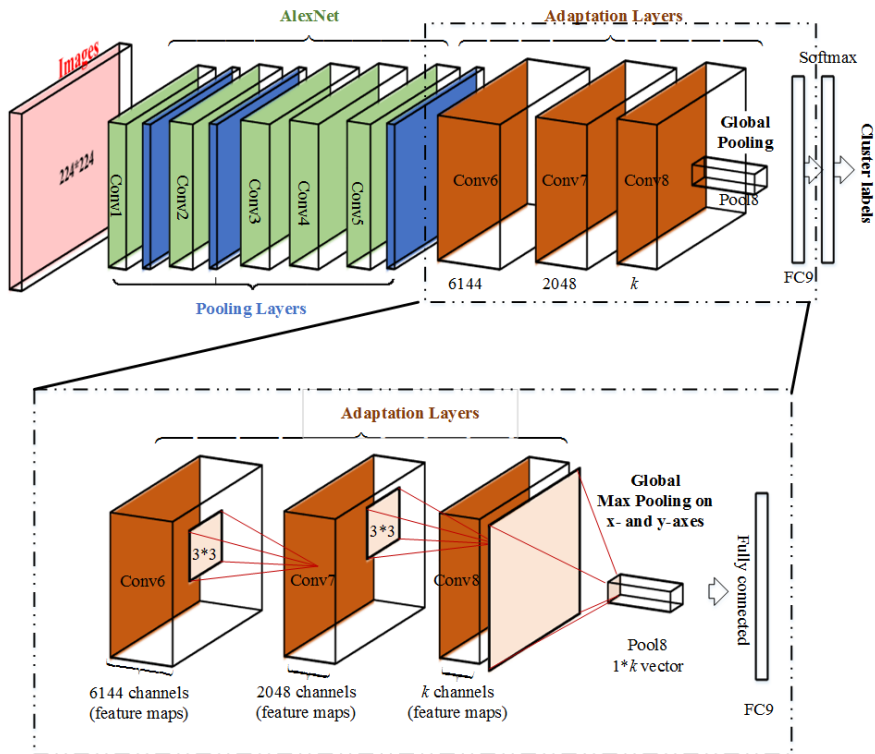

Fig. 1. Network architecture of the proposed FCNN, that is composed of five concatenated convolutional layers (*Conv1-–Conv5*) adopted from AlexNet, followed by three convolutional layers (*Conv6–Conv8*), one fully connected layer (*FC9*) and one softmax layer.

Since the performance of an image clustering method highly relies on the discriminative power of extracted features, there have been quite a few attempts to make use of deep networks to boost image clustering performance by feature representation learning. Nevertheless, the excellent representational power of deep networks relies on a large and comprehensive labeled training dataset, which is not available in unsupervised image clustering tasks. Although a model can be pre-trained based on existing large-scale training image sets, the pre-trained model may not fit the characteristics of input data well.

The first deep-learning-based image clustering work adopts AutoEncoder to learn visual representations followed by conventional *k*-means to obtain final clusters [26]. However, compared to CNN-based architectures, AutoEncoder usually cannot learn representative features well from high-dimensional data such as images. The CNN with Connection Matrix (CNN-CM) method in [27] proposed a connection matrix that allows feeding in additional side information to assist learning discriminative representations for clustering. A full-set *k*-means is then performed to group all images into their corresponding clusters based on the learned features. The complexity of full-set *k*-means will, however, grow drastically when the size of image set becomes large, making large-scale clustering impractical. The CNN with Re-running Clustering (CNN-RC) method in [28] proposed to learn feature representations and cluster images jointly: hierarchical image clustering is performed in the forward pass, while representations are learned in the backward pass. In the hierarchical clustering, image samples are regarded as initial centroids, and then reliable label information is extracted from an undirected affinity/similarity matrix established from the input image set. The network parameters are iteratively updated towards obtaining better feature representations by minimizing a predefined loss metric. Nevertheless, constructing an undirected affinity matrix consumes high computation and memory complexity when the training set becomes large. The memory cost can hardly be reduced since it is not a sparse matrix.

### B. Contribution of Proposed Method

Although CNNs have been shown to achieve good performances in supervised learning-based image/video applications such as visual object localization, tracking, and categorization, existing CNNs cannot well tackle large-scale image clustering as there usually do not exist enough labeled data for feature representation learning of CNNs. To efficiently address the problems of learning representative features from unlabeled input images for large-scale image clustering, we propose a clustering CNN (CCNN) to achieve joint clustering and representation learning. The key idea behind our method is that learning better feature representations of input images leads to better clustering results. Meanwhile, better image clustering will benefit the feature learning of the proposed CNN as well. To reduce computation and memory costs, we incorporate mini-batch *k*-means into the CNN-based clustering framework. The main contribution of this paper is three-fold: i) We are among the first to propose a framework that integrates mini-batch *k*-means with state-of-the-art CNNs to efficiently address the large-scale image clustering problem; ii) we propose a novel iterative centroid updating method that can avoid drift error caused by the feature mismatch between successive iterations of representation learning with mini-batch *k*-means, which was never studied and addressed before; and iii) the proposed framework can be easily integrated into existing CNN-based networks.

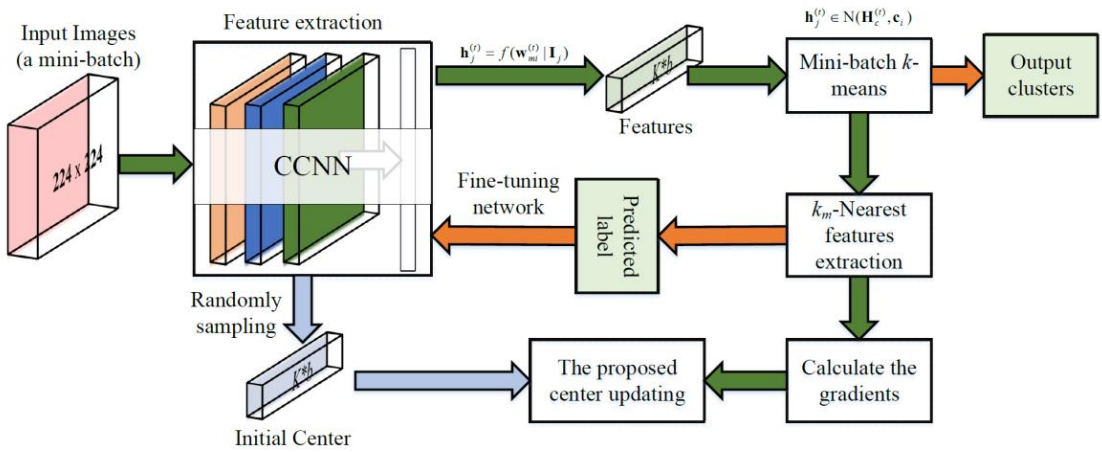

Fig. 2. Block diagram of the proposed CCNN for joint image clustering and representation learning.

The rest of this paper is organized as follows. Sec. II overviews the proposed CCNN architecture. Sec. III presents the proposed joint clustering and representation learning framework. In Sec. IV, experimental results are demonstrated. Finally, Sec. V concludes this paper.

## II. Proposed Clustering CNN Architecture

Most deep-learning-based image clustering approaches estimate the label of an image by passing a whole image through a deep network (e.g., AutoEncoder or CNN) [26]–[28], which tends to extract the global features for the image [19]. Nevertheless, people usually group image clusters according to images' salient features [29]. To extract local salient features from an image, instead of a traditional CNN with several fully connected layers, we propose a CCNN to better capture the salient part of an image without the need of providing any bounding-box in the training stage. As illustrated in Fig. 1, the proposed CCNN is composed of five convolutional layers *Conv1–Conv5* adopted from the first five convolutional layers of AlexNet [19], followed by three adaptation layers (*Conv6–8*) with channel numbers 6144, 2048, and *k*, respectively, that replace the fully connected layers in AlexNet. The adaptation layers involve three convolutional layers, *Conv8−Conv8*, all with 3 × 3 kernels followed by a global max-pooling that finds the maximum value for each channel of *Conv8* so that the size of the output of the global max-pooling is 1 × *k*, where *k* is the number of clusters. In the proposed CCNN, the salient region can be roughly localized by *Conv8*, as reported in [29]. As a result, the proposed CCNN extracts features merely from the salient regions of an image. Note, *Conv1–Conv5* of CCNN can also be replaced with other stacked convolutional layers adopted from ResNet [21], VGG [22], or Inception modules [23] to build a more effective CCNN.

As shown in Fig. 2, to address the complexity issue in large-scale image clustering, we propose to incorporate mini-batch *k*-means clustering into the proposed CCNN network in which the image clustering and feature learning are jointly solved and updated by mini-batch stochastic gradient descent (SGD), making the large-scale image clustering feasible. In our clustering method, at first *k* samples are randomly picked from input data as initial cluster centroids. We then extract the features of input samples using CCNN. For each mini-batch, we perform mini-batch *k*-means to assign cluster labels to individual input samples based on the extracted features, followed by SGD to update the parameters of CCNN. As a result, new features are extracted based on the updated network parameters and then used to re-cluster the input images. The process is iterated until the clustering result converges to a stable point.

## III. Joint Clustering and Representation Learning Based on Mini-Batch K-Means

Let $\mathbb{I} = \{\mathbf{I}_1, \mathbf{I}_2, ..., \mathbf{I}_{N_x}\}$ denote the input image set containing $N_x$ images. The goal is to group $N_x$ images into $k$ clusters $\mathbf{C} = \{\mathbf{c}_1, \mathbf{c}_2, ..., \mathbf{c}_k\}$. Since, when $N_x$ is large, clustering the whole large image set at one time would lead to high computation and memory costs, we propose to divide the input image set into mini-batches of a small and fixed size, and then perform clustering for individual mini-batches. Given a mini-batch containing $N_m$ images randomly sampled from $\mathbb{I}$, the mini-batch's feature set $\mathbf{H} = \{\mathbf{h}_1, \mathbf{h}_2, ..., \mathbf{h}_{N_m}\}$ is extracted from the *FC9* layer of CCNN using filters $\mathbf{h}_i = f(\mathbf{W}_{FC9}|\mathbf{I}_i)$, where $\mathbf{W}_{FC9}$ represents the set of parameters (weights) of *FC9*.

The proposed scheme for iterative image clustering and representation learning is illustrated in Fig. 2. We first initialize the parameters of the CCNN by a pre-trained model (will be elaborated later) for speeding up the convergence of iterations. We then randomly pick $k$ images $\mathbf{I}_c$ from the input image set and extract their features $\mathbf{H}_c$ using the pre-trained CCNN as the initial cluster centroids $\mathbf{C}$. After the initialization, we divide the input image set into mini-batches, and for the *b*-th mini-batch, perform mini-batch *k*-means [29] to assign cluster labels to features $\mathbf{h}_i^{(b)} = f(\mathbf{W}|\mathbf{I}_i^{(b)}) \in \mathbf{H}(\mathbb{I}^{(b)})$ extracted from individual images of the mini-batch. Based on the assigned labels to the feature set of the *b*-th mini-batch, we can update the parameters of the CCNN using SGD. Then, features $\mathbf{h}_i^{(b)}$ of the $b$-th mini-batch are used to update their corresponding centroids using SGD. Since $\mathbf{W}$ will be updated after each iteration, the extracted feature $\mathbf{h}_i^{(b)}$ will also be updated as well, resulting in a possible mismatch between the features extracted in successive iterations. In this case, the centroid updating based on SGD may become unstable and unpredictable since the feature mismatch will lead to gradient drift error in SGD. To overcome this problem, we analyze the gradient drift error between two successive iterations and compensate for the drift error by tracking backward the features extracted in two successive iterations to ensure their consistency. Finally, the proposed method updates the extracted feature $\mathbf{h}$, centroids $\mathbf{C}$, and parameters $\mathbf{W}$ iteratively to mitigate such drift error. The details of the proposed method are elaborated in the following subsections.

### A. Initialization of CCNN

To accelerate the training process, in the proposed CCNN, since *Conv1–Conv5* are part of AlexNet [19], we directly pre-train the parameters of *Conv1–Conv5* in the AlexNet network based on the ILSVRC12 training set of ImageNet [25]. After the pre-training of *Conv1–Conv5*, we concatenate the remaining layers (i.e., *Conv6–Conv8*, *FC9*, and *Softmax*) of the CCNN with *Conv1–Conv5*. During the pre-training process, data argumentation is used to increase sample diversity. After the initialization, the pre-trained set of parameters is then used as an initial model for all image clustering tasks.

### B. Representation Learning

In this work, we extract local salient features from the output of layer *Conv8* [29], and then feed the features into *FC9* to generate the features for clustering. To learn the parameters of *FC9* and *Softmax* of the proposed CCNN, we use a standard SGD process [31] as illustrated in Fig. 3, where parameter sets $\mathbf{W}_{\text{FC9}} = \{w_{mi}\}$ and $\mathbf{W}_{\text{SMax}} = \{w_{ij}\}$ represent the weights of *FC9* and *Softmax*, respectively. In order to learn the weights of *FC9* and *Softmax*, we first define the following sum of squared errors (SSE) objective function:

$$E = \frac{1}{2}\sum_{j=1}^{k}(y_j - t_j)^2, \quad (1)$$

where $k$ is the number of clusters, $y_j$ is the $j$-th cluster label predicted using the CCNN model, and $t_j$ is the $j$-th cluster label predicted using mini-batch $k$-means that is used as a pseudo ground-truth to guide the update of the CCNN model. Then we calculate the gradients of objective function $E$ with respect to $w_{mi}$ and to $w_{ij}$, respectively. We start with the gradient with respect to $w_{ij}$ by using the chain rule as follows:

$$\frac{\partial E}{\partial w_{ij}} = \frac{\partial E}{\partial y_j} \cdot \frac{\partial y_j}{\partial u_j} \cdot \frac{\partial u_j}{\partial w_{ij}}, \quad (2)$$

where $u_j$ is the activation function of the $j$-th ReLU [35]. The derivative of $E$ with respect to $y_j$ is

$$\frac{\partial E}{\partial y_j} = y_j - t_j, \quad (3)$$

and the derivative of ReLU with respect to its input $u$ is

$$\frac{\partial y_j}{\partial u_j} = \max(y_j, 0). \quad (4)$$

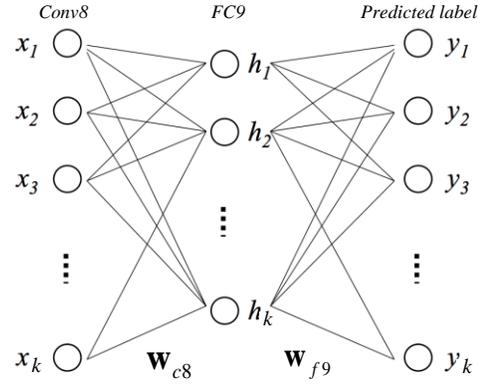

Fig. 3. Illustration of updating the parameters of the *FC9* and *Softmax* layers of CCNN.

The derivative of $u_j = \sum_{i=1}^{k} w_{ij} h_i$ with respect to $w_{ij}$ is

$$\frac{\partial u_j}{\partial w_{ij}} = h_i. \quad (5)$$

As a result, $w_{ij}$ can be updated in the $t$-th iteration by

$$w_{ij}^{(t+1)} = w_{ij}^{(t)} - \eta \cdot (y_j - t_j) \cdot \max(y_j, 0) \cdot h_i, \quad (6)$$

where $\eta$ is the learning rate. Similarly, we can calculate $\frac{\partial E}{\partial w_{mi}}$ by the chain rule:

$$\frac{\partial E}{\partial w_{mi}} = \sum_{j=1}^{k}\left(\frac{\partial E}{\partial y_j} \cdot \frac{\partial y_j}{\partial u_j} \cdot \frac{\partial u_j}{\partial h_i}\right) \cdot \frac{\partial h_i}{\partial u_i} \cdot \frac{\partial u_i}{\partial w_{mi}}, \quad (7)$$

where $\frac{\partial u_j}{\partial h_i} = \frac{\left(\partial \sum_{i=1}^{k} w_{ij} h_i\right)}{(\partial h_i)} = w_{ij}$. The remaining derivative terms of $\frac{\partial E}{\partial w_{mi}}$ include $\frac{\partial h_i}{\partial u_i}$ and $\frac{\partial u_i}{\partial w_{mi}}$, where the derivative of ReLU $h_i$ with respect to $u_i$ is

$$\frac{\partial h_i}{\partial u_i} = \max(h_i, 0), \quad (8)$$

and $\frac{\partial u_i}{\partial w_{mi}}$ is

$$\frac{\partial u_i}{\partial w_{mi}} = \frac{\partial \sum_{m=1}^{k} w_{mi} x_m}{\partial w_{mi}} = x_m. \quad (9)$$

Consequently, $w_{mi}$ can be iteratively updated by

$$\begin{aligned}w_{mi}^{(t+1)} &= w_{mi}^{(t)} - \eta \sum_{j=1}^{k}[(y_j - t_j) \cdot \max(y_j, 0) \cdot w_{ij}] \\ &\quad \cdot \max(h_i, 0) \cdot x_m \\ &= w_{mi}^{(t)} - \eta \Delta w_{mi}^{(t)}\end{aligned} \quad (10)$$

Finally, the full gradient for updating the weights of *FC9* can be calculated by (10).

### C. Cluster Centroid Updating

Assume that the size of a mini-batch is $N_m$, we randomly sample $N_m$ images from the input image set $\mathbb{I}$ to form a mini-batch. Initially, we randomly pick $k$ features $\mathbf{H}_c^{(0)} = \{\mathbf{h}_1^{(0)}, \mathbf{h}_2^{(0)}, \dots, \mathbf{h}_k^{(0)}\}$ from $\mathbb{I}$ as initial centroids $\mathbf{C}$, where $\mathbf{h}_j^{(0)}$ denotes the feature of the $j$-th image in the first iteration (i.e., $t = 0$). To effectively initialize the cluster centroids, we follow the centroid seeds selection strategy proposed in [34] to maximize the diversity among initial centroids. Mini-batch $k$-means is then performed to assign individual samples of each mini-batch to their corresponding clusters. Based on the mini-batch clustering result, the centroids of those clusters that are assigned to the mini-batch's samples are updated based on SGD [31]. In iteration $t$, the $i$-th centroid $\mathbf{c}_i^{(t)}$ that is assigned to a new sample is updated by the weighted average of the features of the $(t-1)$-th centroid and the features of the newly assigned sample $\boldsymbol{h}_{\text{new}}^{(t)}$ as follows:

$$\boldsymbol{c}_i^{(t)} = (1-\gamma_i)\boldsymbol{c}_i^{(t-1)} + \gamma_i \boldsymbol{h}_{\text{new}}^{(t)}, \quad (11)$$

where $\boldsymbol{h}_{\text{new}}^{(t)}$ represents the extracted features of the sample in mini-batch $\mathbf{H}_c$ that is newly assigned to its nearest neighbor centroid $\boldsymbol{c}_i$. We follow [31] to use per-centroid learning rates $\gamma_i$ for the $i$-th centroid as determined by

$$\boldsymbol{\gamma}_i = 1/\text{count}(\boldsymbol{c}_i), \quad (12)$$

where count($\mathbf{c}_i$) is the number of samples assigned to $\mathbf{c}_i$.

### D. Compensation of Feature Drift

Note, in the $t$-th iteration, the feature vector of the $j$-th image $\mathbf{h}_j^{(t)} = f(\mathbf{W}_{\text{FC9}}^{(t)}|\mathbf{I}_j)$ is extracted based on the filter coefficients $w_{mi}^{(t)}$ of FC9. However, $w_{mi}^{(t)}$ is updated along time during representation learning, thereby making $\mathbf{h}_j^{(t)}$ vary along time as well. The time-varying nature of $\mathbf{h}_j^{(t)}$ leads to the inconsistency between the features extracted from the same image in two successive iterations. For example, $\mathbf{h}_j^{(t)} = f(\mathbf{W}_{\text{FC9}}^{(t)}|\mathbf{I}_j)$ extracted in iteration $t$ is different from $\mathbf{h}_j^{(t-1)} = f(\mathbf{W}_{\text{FC9}}^{(t-1)}|\mathbf{I}_j)$ in iteration $t-1$, as $\mathbf{W}_{\text{FC9}}^{(t)}$ and $\mathbf{W}_{\text{FC9}}^{(t-1)}$ are different due to parameter updating. This makes centroid updating in (11) unreliable since $\boldsymbol{h}_j^{(t)}$ is time varying, which can significantly degrade the performance of image clustering as will be demonstrated in Sec. IV. To address this mismatch problem, we propose an approach to ensure feature consistency between two successive iterations. In iteration $t$, we have $\mathbf{c}_i^{(t)} = (1-\gamma_i)\mathbf{c}_i^{(t-1)} + \gamma_i \mathbf{h}_j^{(t)}$ and $w_{mi}^{(t)} = w_{mi}^{(t-1)} - \eta \Delta w_{mi}^{(t-1)}, \forall m, i$. As a result, the feature extracted in the $(t-1)$-th iteration can be backward tracked from the weight obtained in iteration $t$ by

$$\mathbf{h}_j^{(t-1)} = f\left((\mathbf{w}_{\text{FC9}}^{(t)} + \eta \Delta \mathbf{w}_{\text{FC9}}^{(t-1)})|\mathbf{I}_j\right). \quad (13)$$

To maintain the consistency between the features used in two successive iterations, we replace the features $\boldsymbol{h}_{\text{new}}^{(t)}$ in (11) with the backward tracked features in (13), and reformulate the centroid updating as follows:

$$\mathbf{c}_i^{(t)} = (1-\gamma_i)\mathbf{c}_i^{(t-1)} + \gamma_i f\left((\mathbf{w}_{\text{FC9}}^{(t)} + \eta \Delta \mathbf{w}_{\text{FC9}}^{(t-1)})|I_j\right). \quad (14)$$

In this way, the cluster centroids can be properly updated. After iterating for several epochs with the proposed framework, the cluster labels of images will converge to their final values more reliably. Besides, the iteratively fine-tuned network parameters can be used to extract successively improved visual representations for image clustering. Furthermore, the proposed mini-batch-based scheme can deal with large-scale image clustering on a single personal computer with reasonable

Table I.
PROPOSED JOINT REPRESENTATION LEARNING AND CLUSTERING ALGORITHM

| | |
|---|---|
| 1. | Given: $k$, mini-batch size $N_m$, max iteration no. $T$, Image dataset $\mathbb{I}$ |
| 2. | Randomly sample $k$ samples from $\mathbb{I}$ as entroids $\mathbf{c}_i \in \mathbf{C}$ |
| 3. | Extract image features from the initial centroids |
| 4. | v ← 0 |
| 5. | For $t = 1$ to $T$ do: |
| 6. |    $\mathbf{M} \leftarrow N_m$ images features picked randomly from $\mathbb{I}$ |
| 7. |    For $\mathbf{m} \in \mathbf{M}$ do: |
| 8. |      $\mathbf{y}, \gamma(\mathbf{m}), \mathbf{d} = N(\mathbf{m}, \mathbf{C})$ |
| 9. |      //find label **y**, distance **d**, and learning rate |
| 10. |    End For |
| 11. |    $\mathbf{M}', \mathbf{y}' \leftarrow k$-NN($\mathbf{d}, \mathbf{M}, \mathbf{C}, k_m$) |
| 12. |    // Assign top-$k_m$ samples & labels to set **M'** & **y'** |
| 13. |    If (size(**M'**) = $b$) fo |
| 14. |      Finetune(CCNN, **M'**, **y'**, $k_m$) |
| 16. |      //Use the predicted labels to fine tune CCNN |
| 17. |      For $\mathbf{m} \in \mathbf{M}$ do: |
| 18. |         Update the centroid by (14) //Adaptive centroid updating |
| 19. |      End For |
| 20. |    End If |
| 21. | End For |

computational and memory complexity as will be shown in the experiment section.

### E. Top-$k_m$ Based Parameter Updating

Since the predicted cluster labels of the samples in a mini-batch may not be all reliable because the network parameters of CCNN may be inaccurate, we only pick from a mini-batch the top-$k_m$ samples with the smallest distances to their corresponding centroids to update the network parameters of CCNN. In this way, we update the parameters of CCNN once when collecting $N_m$ samples from every $N_m/k_m$ times of mini-batch clustering, where $N_m$ is the size of a mini-batch. Note, the higher the $k_m$ value is, the faster the parameter updating process will be, but the lower the performance of clustering due to the lower representational power of the parameters of CCNN. In contrast, a much smaller $k_m$ value, though achieving better clustering performance, would result in a drastically increased number of updating processes and long training time. In our experiments, $k_m$ is empirically set to be 10. The proposed algorithm is summarized in TABLE I.

## IV. EXPERIMENTAL RESULTS

### A. Experiment Setup

*1) Comparison Schemes*: To evaluate the performance of the proposed method, we test our method against three state-of-the-art deep-learning-based image clustering schemes including the AutoEncoder-based Deep Embedding Clustering (DEC) scheme proposed in [26], the CNN with Connection Matrix (CNN-CM) method proposed in [27], and the CNN with Re-running Clustering (CNN-RC) [28]. Note, as explained above, these three deep-learning-based schemes cannot deal with large-scale image sets consisting of millions of images on a personal computer equipped with a commercial GPU(s) like Titan X. Therefore, besides the three methods, we also implemented three baseline schemes for performance evaluation: 1) Baseline-I: the proposed method without feature mismatch compensation, that is, using (11) instead of (14) to update cluster centroids; 2) Baseline-II: mini-batch k-means clustering based on the pre-trained model described in Sec. 3A without iterative representation learning; 3) Baseline-III: full-set k-means clustering based on the pre-trained model described in Sec. 3A without iterative representation learning.

Table II.
NMI Performance Comparison between The Proposed Scheme and State-of-The-Art Schemes for Three Image Datasets.

| Evaluated methods | ILSVRC-Val | Places-Val | Places-Train |
|---|---|---|---|
| DEC [26] | 0.155 | 0.113 | N.A. |
| CNN-CM [27] | 0.137 (Ran.)<br>0.225 (Pre.) | 0.198 (Ran.)<br>0.237 (Pre.) | N.A. |
| CNN-RC [28] | 0.295 (Ran.)<br>0.369 (Pre.) | 0.213 (Ran.)<br>**0.310 (Pre.)** | N.A. |
| Baseline-I | 0.181 (Pre.) | 0.153 (Pre.) | 0.047 (Pre.) |
| Baseline-II | 0.231 (Pre.) | 0.177 (Pre.) | 0.045 (Pre.) |
| Baseline-III | 0.293 (Pre.) | 0.201 (Pre.) | N.A. |
| **Proposed** | 0.314 (Ran.)<br>**0.375 (Pre.)** | 0.219 (Ran.)<br>0.307 (Pre.) | 0.166 (Ran.)<br>**0.187 (Pre.)** |

Table III.
Comparison of Run-time and Memory Costs between The Proposed Scheme and State-of-The-Art Schemes for Three Image Datasets.

| Evaluated methods | ILSVRC-Val | Places-Val | Places-Train |
|---|---|---|---|
| DEC [26] | 0.9 hr/16 GB | 0.75 hr/14 GB | N.A. |
| CNN-CM [27] | 3 hr/ 7 GB | 1.8 hr/5 GB | N.A. |
| CNN-RC [28] | 5.1 hr/10 GB | 4.6 hr/7 GB | N.A. |
| Baseline-I | 1.1 hr/8 GB | 0.5 hr/8 GB | 40 hr/8 GB |
| Baseline-II | 0.9 hr/22 GB | 0.45 hr/19 GB | **36 hr/8 GB** |
| Baseline-III | 4.28 hr/22 GB | 3.68 hr/19 GB | N.A. |
| **Proposed** | **1.2 hr/8 GB** | **0.5 hr/8 GB** | 43 hr/8 GB |

2) *Datasets for Pre-training and Testing*: We selected two large-scale image datasets, ILSVRC12 in ImageNet [25] and Places2 [32], for clustering performance evaluation. ILSVRC12 consists of 1.2 million training images and 50,000 validation images collected from 1,000 object categories, and Places2 consists of 1.6 million training images and 18,250 validation images collected from 356 scene categories. Since, for fast convergence, the parameters of CCNN were pre-trained from the ILSVRC12 training set (denoted "ILSVRC-Train"), we did not evaluate the performances of the clustering methods on the ILSVRC12 training set for fairness. Instead, we conducted performance evaluation on the Places2 training (denoted "Places-Train") and validation (denoted "Places-Val") sets, and also on the ILSVRC12 validation set (denoted "ILSVRC-Val"). For the Places2 training and validation sets, the channel number to *Conv8* and the number of neurons of *Softmax* in the proposed CCNN were both set to 365, whereas for the ILSVRC12 validation set the number of channels to *Conv8* and number of neurons of *Softmax* were both set to 1000. Similar to [7], all test images were cropped to $256 \times 256$ center-surrounding images.

Besides the large-scale datasets, we also evaluated the performances of the clustering methods on a smaller scale image dataset MNIST, which contains 60,000 greyscale images of size 28x28. Since the size of MNIST handwriting images is small, the images were not cropped.

3) *Computation Platform*: We implemented the proposed method on top of TensorFlow [36] on an Intel Core i7-4770 PC with 32 GB RAM which is equipped with an NVIDIA Titan X GPU with 12 GB GPU RAM.

B. *Performance Evaluation*

To evaluate the objective clustering performances of the proposed method and the compared methods, we adopt the widely used metric: Normalized Mutual Information (NMI) [33] as defined below:

$$\text{NMI}(\mathbf{t}, \mathbf{y}) = \frac{I(\mathbf{t}, \mathbf{y})}{\sqrt{H(\mathbf{t})H(\mathbf{y})}}, \quad (15)$$

where $H(\cdot)$ stands for the entropy, and $I(\mathbf{t}, \mathbf{y}) = H(\mathbf{t}) - H(\mathbf{t}|\mathbf{y})$ denotes the mutual information. The higher the NMI is, the more reliable the clustering result becomes.

Table II compares the NMI performances of the proposed method, DEC [26], CNN-CM [27], CNN-RC [28] and the three baseline methods for three image sets. For the CNN-based methods including the proposed CCNN, CNN-CM, and CNN-RC, we compare the performances of these methods with a pre-trained model learned from the ILSVRC-Train set (denoted "Pre.") and with random initialization (denoted "Ran."). As for the three baseline methods, we only compare the performances with a pre-trained model. As shown in Table II, with the pre-trained model, the proposed method achieves comparable NMI performances with CNN-RC and significantly outperforms CNN-CM and DCC for ILSVRC-Val and Places-Val. As for the large-scale image dataset Places-Train which contains millions of images, only the proposed CCNN can successfully cluster such a large-scale image dataset on a personal computer equipped with a commercial GPU card, whereas DEC, CNN-CM, and CNN-RC, and Baseline-III all cannot handle large-scale image clustering due to their high complexity as will be explained later. Compared with Baseline-I, we can observe that the feature mismatch in mini-batch-based centroid updating leads to significant drifting error which degrades the NMI performance by 0.14–0.19. Compared with the direct combination of a pre-trained model with mini-batch k-means (Baseline-II) and full-set *k*-means (Baseline-III), the proposed joint optimization of clustering and parameter learning leads to performance improvement in NMI by 0.13–0.14 and 0.08–0.10, respectively.

Table III compares the memory and run-time costs for three image sets, where we set the number of epochs for parameter updating to 10. The run-time is proportional to the size of image set and the number of clustering iterations. The comparison shows that, for mini-batch size $N_m = 50$, our method consumed about 8GB GPU memory and 43 hours to obtain the clustering result of the Places-Train image set on Titan X, whereas DEC, CNN-CM, and CNN-RC all failed in this clustering task. DEC [26] learns feature representations from a training set based on AutoEncoder. However, it has been shown that the representation learning performance of an AutoEncoder-based network is generally unsatisfactory for high-dimensional data (e.g., images) in terms of computation and clustering performance [7]. Although CNN-based networks have proven to achieve good representational power, both CNN-CM [27] and Baseline-III perform full-set k-means clustering which needs to extract the features of all images and compare the distances between features, leading to huge computation/memory requirement and making large-scale clustering infeasible on a single general-purpose PC equipped with a GPU graphic card. Similarly, CNN-RC [28] relies on constructing an $N_x \times N_x$ affinity matrix, making the clustering process unsolvable when the size of dataset $N_x$ is large. Instead of using computation/memory demanding operations like full-set k-means and affinity matrix construction, the proposed mini-batch-based method with feature drift compensation can efficiently and reliably address the problem of large-scale joint representation learning and clustering.

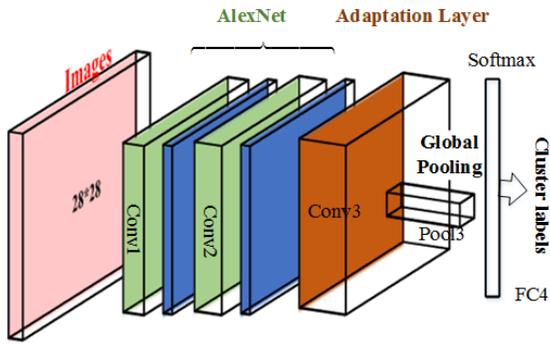

Fig. 4. Simplified CCNN for the MNIST dataset.

Table IV.
NMI PERFORMANCE COMPARISON OF THE PROPOSED METHOD, K-MEANS, AND CNN-SF/CNN-RC FOR MNIST IMAGE SETS.

| Evaluated methods | MNIST-Train | MINST-Test |
|---|---|---|
| k-means | 0.500 | 0.528 |
| CNN-SF [28] | 0.906 | 0.876 |
| CNN-RC [28] | **0.913** | 0.915 |
| Proposed | 0.876 | **0.916** |

Table V.
NMI PERFORMANCE COMPARISON OF DIFFERENT CLUSTERING METHODS ON TOP OF THE NETWORK ARCHITECTURES OF CCNN AND ALEXNET [7].

| Evaluated methods | ILSVRC-Val | Places-Val |
|---|---|---|
| Baseline-II with AlexNet | 0.220 | 0.168 |
| Baseline-II with CCNN | 0.231 | 0.177 |
| Baseline-III with AlexNet | 0.279 | 0.194 |
| Baseline-III with CCNN | 0.293 | 0.201 |
| Proposed with AlexNet | 0.346 | 0.291 |
| Proposed with CCNN | 0.375 | 0.307 |

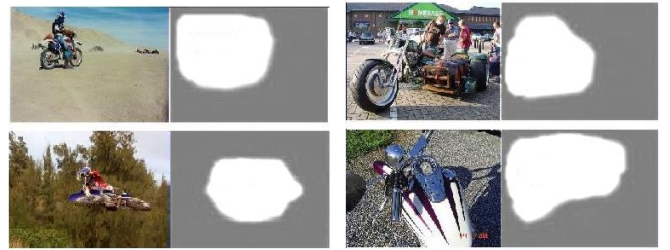

Fig. 5. Visualized feature maps of *Conv8* using global average pooling showing that the objects (motorcycles) can be roughly localized.

We also evaluated the effectiveness of the proposed method on the MNIST dataset. Because the image resolution of MNIST is much smaller than that of ImageNet, we built a simplified version of CCNN by removing some convolution layers to fit the data type, as depicted in Fig. 4. As shown in TABLE IV, the performance of the proposed method significantly outperforms k-means. Compared to the state-of-the-art CNN-based methods CNN-SF/CNN-RC in [28], where CNN-SF is a simplified version of CNN-RC, our method achieves comparable performance on MNIST. More results about the comparison with other typical clustering methods can be found in [28] which show that CNN-SF/CNN-RC outperformed many other schemes. All the results show that the proposed CCNN performs well for image datasets of various scales.

As suggested in [29], localizing salient objects in an image can benefit the categorization of the image. To this end, the network architecture of CCNN is mainly modified from AlexNet [7] by replacing the fully-connected layers of AlexNet with the adaptation layer that consists of three convolutional layers (*Conv5–Conv8*). To evaluate the effectiveness of the proposed network structure of CCNN, we compare the performances of three clustering schemes: the project joint clustering and parameter updating, and the Baseline II and Baseline III schemes described in Sec. IV.A on top of the network architectures of CCNN and AlexNet on the ILSVRC-Val and Places-Val datasets. As shown in TABLE V, the proposed CCNN architecture achieves better performances for all the three clustering schemes compared to AlexNet. As shown in Fig. 5, the visualized feature maps of *Conv8* using global average pooling all illustrate that the proposed architecture can roughly localize salient regions, making the following layers (i.e., *FC9* and *Softmax*) learn the feature representations from salient regions only. It is one reason that the proposed method outperforms the others in image clustering.

### C. Impacts of $k_m$ and the number of Epochs

As mentioned in Sec. III.E, we only pick top $k_m$ samples to update the network parameters of CCNN in each mini-batch. Fig. 6 shows the impact of different $k_m$ values on the clustering performance for the ImageNet validation set (ILSVRC-Val). It shows that the $k_m = 1$ achieves the best NMI performance, which, however, consumes the longest fine-tuning time for learning the network parameters. To achieve a good tradeoff between performance and complexity, the combination of $N_m = 50$ and $k_m = 10$ seems to be a reasonable choice that leads to a slightly lower NMI performance but much fewer computation compared with $k_m = 1$. Note, when the mini-batch size equals to the size of image set and $k_m = 1$, this extreme case is similar to CNN-RC [28] but in a mini-batch optimization form. Compared to CNN-RC, the main advantage of CCNN is that the mini-batch optimization form can deal with large-scale image clustering problems.

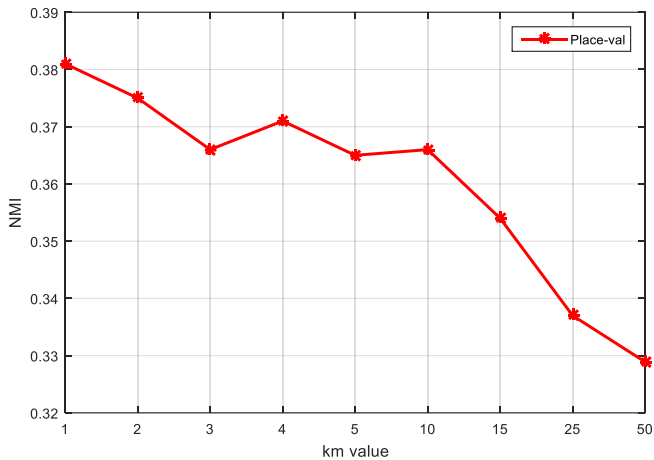

Fig. 6. Comparison of NMI performance of the proposed method with different $k_m$ settings for the ILSVRC-Val dataset.

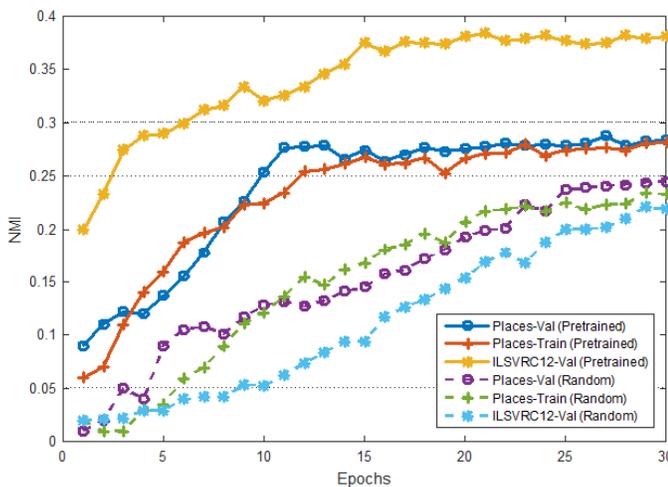

Fig. 7. Comparison of NMI performances of the proposed approach with random initial and pre-trained model for three test datasets (ILSVRC12-Val, Places-Train, and Places-Val) versus the number of training epochs.

Since the clustering performance and computational complexity of our mini-batch-based CCNN basically increase with the number of epochs for parameters updating, we also evaluated the performance of CCNN on the three image datasets with different numbers of epochs. Fig. 7 compares the proposed approach with random initialization and with a pre-trained model for three test data sets (ILSVRC12-Val, Places-Train, and Places-Val) versus the number of epochs of training. As illustrated in Fig. 7, the NMI performance of the CCNN with a pre-trained model becomes saturated after 14 epochs for all test image sets, whereas that of the CCNN with random initialization becomes saturated after about 30 epochs. With $N_m = 50$ and $k_m = 10$, all samples in the input image set can be picked at least one time every five epochs of the iterative updating process. We therefore suggest setting the number of epochs to be 10 or less than 10 for the parameter fine-tuning process of CCNN to achieve a good tradeoff between clustering performance and computational complexity.

### D. Convergence Analysis of CCNN

Although k-means clustering is guaranteed to converge, there is no theoretical guarantee for the convergence of mini-batch k-means based clustering approaches. Nevertheless, many recent studies in deep CNN models based on mini-batch training have shown that, with a reasonable initial model pre-trained from a comprehensive data set (e.g., ImageNet), deep CNNs can usually converge reliably in the training process. Since our mini-batch *k*-means clustering method adopts an initial model pre-trained from ImageNet to estimate cluster labels, it can usually lead to reliable convergence performance. To evaluate the convergence performance of our method with a pre-trained model, in our experiments shown in Fig. 7, we adopt a pre-trained model learned from ImageNet and then use it to cluster images of a different dataset (e.g., Places2) for fair comparison. Besides, we also conduct experiments to evaluate the convergence performance of our method with random initialization empirically. Fig. 7 shows that the clustering performances of our approach with random initialization of parameters for the three test datasets are lower than that with a pre-trained model. However, we can also observe that the NMI performances of CCNN with a pre-trained model and with a random initial model both increase with the number of training epochs, implying that both models make CCNN converge steadily. Therefore, although there is no theoretical guarantee of convergence (same with other existing deep CNN models based on mini-batch training) for a large-scale dataset, our experiments show that our method can achieve a reasonable convergence performance with a pre-trained model or with random initialization.

### 5. CONCLUSION

In this paper, we proposed a clustering convolution neural network (CCNN) architecture, which can extract salient features to benefit image clustering. On top of CCNN, we also proposed a mini-batch-based iterative representation learning and cluster centroid updating approach for efficient large-scale image clustering involving up to millions of images at reasonable memory and computation costs. While the mini-batch iterative updating strategy offers good scalability to the proposed CCNN, we have also proposed a feature drift compensation scheme to avoid the performance degradation due to feature drifting in the mini-batch based iterative process. Our experimental results demonstrate the superior performance and scalability of our method on several public image datasets.

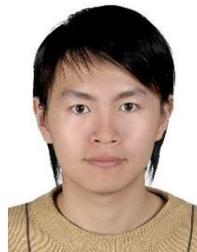

Chih-Chung Hsu received his B.S. degree in Information management from Ling-Tung University of Science and Technology, Taiwan, in 2004, and M.S. and Ph.D. degrees in Electrical Engineering from National Yunlin University of Science and Technology and National Tsing Hua University (NTHU), Taiwan, in 2007 and 2014, respectively.

Dr. Hsu is a postdoctoral researcher with the Institute of Communications Engineering, NTHU. His research interests mainly lie in computer vision, and image and video processing. He received a top 10% paper award from IEEE MMSP 2013.

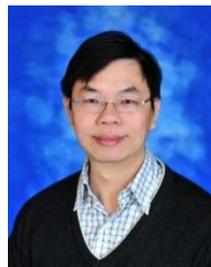

Chia-Wen Lin (S'94-M'00-SM'04) received his Ph.D. degree in electrical engineering from National Tsing Hua University (NTHU), Hsinchu, Taiwan, in 2000.

He is currently a Professor with the Department of Electrical Engineering and the Institute of Communications Engineering of NTHU. He is also Director of the Multimedia Technology Research Center, College of Electrical Engineering and Computer Science of NTHU, Taiwan. He was with the Department of Computer Science and Information Engineering, National Chung Cheng University, Chiayi, Taiwan, during 2000–2007. Prior to that, he worked for the Information and Communications Research Labs, Industrial Technology Research Institute, Hsinchu, Taiwan, during 1992–2000, where his final post was Section Manager. His research interests include image/video processing and video networking.

Dr. Lin has served as an Associate Editor of the IEEE TRANSACTION ON IMAGE PROCESSING, the IEEE TRANSACTIONS ON MULTIMEDIA, the IEEE TRANSACTIONS ON CIRCUITS AND SYSTEMS FOR VIDEO TECHNOLOGY, the IEEE MULTIMEDIA, and the *Journal of Visual Communication and Image Representation*. He also served as a Steering Committee member of the IEEE Transactions on Multimedia. He was Chair of the Multimedia Systems and Applications Technical Committee of the IEEE Circuits and Systems Society. He served as Technical Program Co-Chair of the IEEE ICME in 2010, and Special Session Co-Chair of IEEE ICME in 2009. His papers won the Best Paper Award of IEEE VCIP 2015, and the Young Investigator Award of VCIP 2005. He received the Young Investigator Awards presented by Ministry of Science and Technology, Taiwan, in 2006.